# Two-step Authentication: Multi-biometric System Using Voice and Facial Recognition

*Kuan Wei Chen[1], Ting Yi Lin[1], Wen Ren Yang[1], Aryan Kesarwani[2], Riya Singh[2]*

[1]*Electronical Engineering, National Changhua University Of Education, Changhua, Taiwan*
[2]*Apex Institute of Technology, Chandigarh University, Gharuan, Punjab, India*
elaping5691@gmail.com, brian104120@gmail.com, wry87c@cc.ncue.edu.tw,
kesarwaniaryan76@gmail.com, riya0502singh@gmail.com,

**Abstract**
The proposed study presents an authentication system integrating facial and voice recognition technologies to enhance authentication in a small group with camera and microphone, providing a solution focused on cost-effectiveness. Through practical implementation and rigorous testing, this project demonstrates the effective integration of diverse biometric modalities, offering security and usability. The face recognition model and voice recognition model serve an efficient accuracy. The voice verification process can contain a real-time authentication procedure that aids in double authentication security by employing random words as CAPTCHAs in the future. The code and models are publicly available at https://github.com/NCUE-EE-AIAL/Two-step-Authentication-Multi-biometric-System.

## 1. Introduction

Unlike previous research aiming to enhance accuracy in multi-biometric systems [1], this study focuses on reducing hardware costs via a two-step biometric process. Highlighting the advantages of multimodal systems [2, 3], our decision-level fusion system leverages built-in cameras and microphones in mobiles and laptops. This approach provides a cost-effective and robust solution, inspired by [4].

Fig 1 shows the workflow of the system, beginning with facial recognition. This simplifies the subsequent voice authentication process, as the voice recognition model only needs to verify the voice in the identified individual. This approach allows our system to efficiently and accurately identify individuals in a small group.

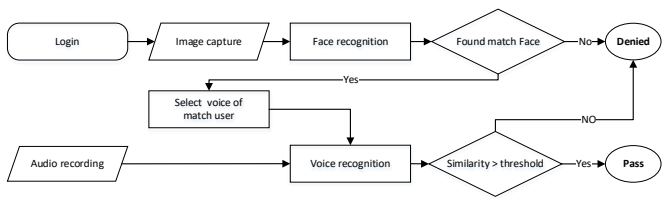

Fig.1 System workflow

## 2. Methodology

*2.1 Dataset*
Dataset containing images of five individuals was collected. Several data augmentation techniques were used to the original photos in order to improve the machine learning model's robustness. The model's capacity to generalize to new data was enhanced by these augmentation techniques, which increased the size of the initial dataset to a final count of 924 photos. In order to reduce background noise and concentrate the dataset on pertinent parts, the MTCNN algorithm was used to identify and clip facial regions from the images. The LibriSpeech dataset was employed for evaluating the performance of the voice model.[5] The training process used the subsets train-other-360, including 921 speakers. For validation, the subsets test-clean with 40 speakers was used.

*2.2 Neural Network Architecture*
Table 1 below shows the voice model architecture, which revolves around convolutional neural network (CNN). A 5x5 convolutional filter with 2x2 stride is applied, followed by a residual block consisting of three convolutional layers: 1x1, 3x3, and 1x1, all with a 1x1 stride. After the mean and affine layer, the triplet loss is applied with 512 dimensions.[6]

Table 1 Architecture of voice recognition model

| Layer | Structure | Stride | Params |
|---|---|---|---|
| Conv64 | $5 \times 5, 64$ | $2 \times 2$ | 1.9K |
| Res64 | $\begin{bmatrix} 1 \times 1, 64 \\ 3 \times 3, 64 \\ 1 \times 1, 64 \end{bmatrix} \times 3$ | $1 \times 1$ | $46K \times 3$ |
| Conv128 | $5 \times 5, 128$ | $2 \times 2$ | 205K |
| Res128 | $\begin{bmatrix} 1 \times 1, 128 \\ 3 \times 3, 128 \\ 1 \times 1, 128 \end{bmatrix} \times 3$ | $1 \times 1$ | $228K \times 3$ |
| Conv256 | $5 \times 5, 256$ | $2 \times 2$ | 821K |
| Res256 | $\begin{bmatrix} 1 \times 1, 256 \\ 1 \times 3, 256 \\ 1 \times 1, 256 \end{bmatrix} \times 3$ | $1 \times 1$ | $953K \times 3$ |
| Conv512 | $5 \times 5, 512$ | $2 \times 2$ | 3.28M |
| Res512 | $\begin{bmatrix} 1 \times 1, 512 \\ 1 \times 3, 512 \\ 1 \times 1, 512 \end{bmatrix} \times 3$ | $1 \times 1$ | $2.89M \times 3$ |
| mean | - | - | 0 |
| affine | $2048 \times 512$ | - | 1M |
| triplet | - | - | 0 |
| total | - | - | 16.8M |

The proposed face recognition model is presented in Fig 2, based on the VGG-16 architecture. The top and dense layer were created with custom layer designed to classify the five individuals in the dataset by learning and distinguishing their unique facial features. The data layers were thoroughly pruned with the objective to derive the optimal accuracy. In order to accommodate photos downsized to 224x224 pixels, the input layer was adjusted. Because the architecture used fewer resources and produced accurate results, it was viable.

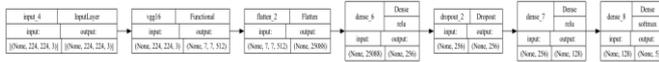

Fig. 2 Architecture of face recognition model

## 3. Result

*3.1 Face Recognition*
Table 2 shows the results of VGG-16 achieved by running multiple epochs of the face recognition model, detailed in the confusion matrix in Fig 3. To evaluate the performance of face recognition model, Fig 4 is showed below.

Table 2 Face recognition results

| System | Accuracy | F1-Score | Recall | Precision |
|---|---|---|---|---|
| VGG16 | 95.135% | 95.732% | 95.153% | 96.317% |

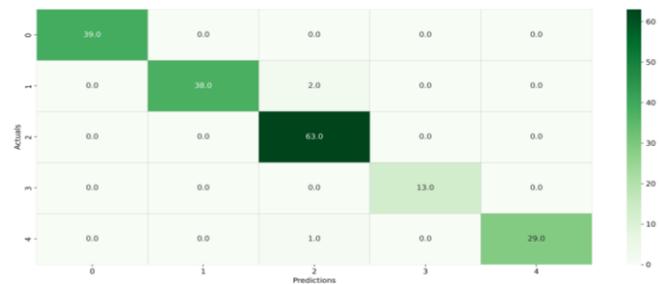

Fig. 3 Confusion matrix of face recognition model

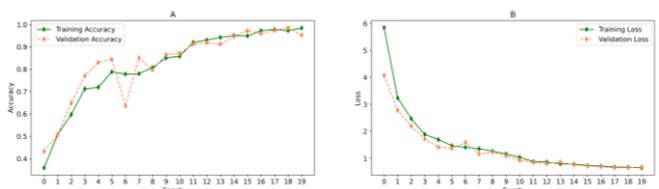

Fig. 4 Graph for epochs of face recognition model showing (a) Training and validation accuracy (b) Training and validation loss.

*3.2 Voice Recognition*
The audio is processed with energy-based voice activity detection to remove silence part. Features are then extracted into 64-dimensional Filterbank coefficients. Table 3 presents the result of voice recognition model after 60 epochs training process. The accuracy and equal error rate (EER) demonstrate the outstanding performance, while the recall and precision highlight its stability. The Adam optimizer was used with default parameters, and the model utilizes triplet loss with minibatch 32 and 0.1 margin α. Fig 5 is to show the evaluation result of voice recognition model.

Table 3. Voice recognition results

| System | Accuracy | EER | Recall | Precision |
|---|---|---|---|---|
| model | 98.9% | 3.456% | 86.48% | 88.65% |

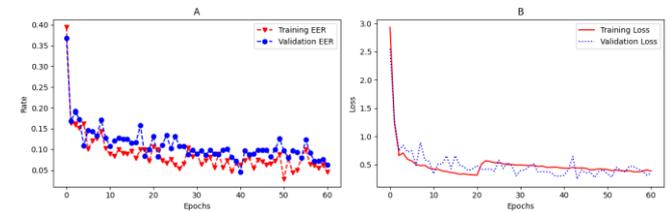

Fig. 5 Graph for epochs of voice recognition model showing (a) Training and validation EER (b) Training and validation loss.

## 4. Conclusion

The multi-model biometric authentication system integrates facial and voice recognition technologies. Leveraging device cameras and microphones, it offers a cost-effective solution. The face recognition model achieves an accuracy of 95.1%, while the voice recognition model achieves an equal error rate of 3.456%. These results underscore the effectiveness of our multimodal biometric authentication system in providing robust and reliable identification capabilities.

Furthermore, by using random sentences as CAPTCHAs in the voice verification process or CMBA system features a real-time authentication process mention in [2], the system can prevent machine attacks, making it practical for implementation on universal devices.

## 5. Acknowledgements